\documentclass[sn-apa]{sn-jnl}% APA Reference Style 
\usepackage{graphicx}%
\usepackage{multirow}%
\usepackage{amsmath,amssymb,amsfonts}%
\usepackage{amsthm}%
\usepackage{mathrsfs}%
\usepackage[title]{appendix}%
\usepackage{xcolor}%
\usepackage{textcomp}%
\usepackage{manyfoot}%
\usepackage{booktabs}%
\usepackage{algorithm}%
\usepackage{algorithmicx}%
\usepackage{algpseudocode}%
\usepackage{listings}%
\usepackage{etoolbox}
\usepackage{outlines}

\usepackage{blindtext}
\usepackage{geometry}
\theoremstyle{thmstyleone}%
\theoremstyle{thmstyletwo}%

\theoremstyle{thmstylethree}%

\begin{document}

\title[GPT-4V]{NERIF: GPT-4V for Automatic Scoring of Drawn Models}


\author[1,2]{\fnm{Gyeong-Geon} \sur{Lee}}\email{ggleeinga@uga.edu}
\author*[1,2]{\fnm{Xiaoming} \sur{Zhai}}\email{xiaoming.zhai@uga.edu}

% \author[1,2]{\fnm{Ehsan} \sur{Latif}}\email{ehsan.latif@uga.edu}
% \author[5,2]{\fnm{Gengchen} \sur{Mai}}\email{gengchen.mai25@uga.edu}

% \author[1]{\fnm{Matthew} \sur{Nyaaba}}\email{Matthew.Nyaaba@uga.edu}

% \author[2]{\fnm{Xuansheng} \sur{Wu}}\email{xuansheng.wu@uga.edu}

% \author[2]{\fnm{Ninghao} \sur{Liu}}\email{ninghao.liu@uga.edu}

% \author[1,3]{\fnm{Guoyu} \sur{Lu}}\email{guoyu.lu@uga.edu}

% \author[4]{\fnm{Sheng} \sur{Li}}\email{vga8uf@virginia.edu}

% \author[1,2]{\fnm{Tianming} \sur{Liu}}\email{tliu@uga.edu}

% \author[6]{\fnm{Cassie Chen} \sur{Cao}}\email{ccao@carnegielearning.com}

% \author*[1]{\fnm{Xiaoming} \sur{Zhai}}\email{xiaoming.zhai@uga.edu}

\affil[1]{\orgdiv{AI4STEM Education Center}, \orgname{University of Georgia}, \orgaddress{\city{Athens}, \state{GA}, \country{United States}}}

\affil[2]{\orgdiv{Department of Mathematics, Science, and Social Studies Education}, \orgname{University of Georgia}, \orgaddress{\city{Athens}, \state{GA}, \country{United States}}}
% \street{110 Carlton St.}, 

% \street{Elson 178}, 

\pretocmd{\abstractname}{}{}{}
\abstract{
Engaging students in scientific modeling practice in the classroom is critical to improving students' competence in using scientific knowledge to explain phenomena or design solutions. However, scoring student-drawn models is time-consuming. The recently released GPT-4V provides a unique opportunity to advance scientific modeling practices by leveraging the powerful image classification capability. To test this ability specifically for automatic scoring, we developed a method NERIF (Notation-Enhanced Rubric Instruction for Few-shot Learning) employing \textit{instructional note} and rubrics to prompt GPT-4V to score students' drawn models for science phenomena. We randomly selected a set of balanced data (N = 900) from a parental study that includes student-drawn models for six modeling assessment tasks. Each model received a score from GPT-4V ranging at three levels: 'Beginning,' 'Developing,' or 'Proficient' according to scoring rubrics. GPT-4V scores were compared with human experts' consent scores to calculate scoring accuracy. Results show that GPT-4V's average scoring accuracy was mean =.51, SD = .037, with varying accuracy across scoring categories. Specifically, average scoring accuracy was .64 for the 'Beginning' class, .62 for the 'Developing' class, and .26 for the 'Proficient' class, indicating that more proficient models are more challenging to score. Further qualitative study reveals how GPT-4V retrieves information from image input, including problem context, example evaluations provided by human coders, and students' drawing models. We also uncovered how GPT-4V catches the characteristics of student-drawn models and narrates them in natural language. At last, we demonstrated how GPT-4V assigns scores to student-drawn models according to the given scoring rubric and instructional notes. Our findings suggest that the NERIF method is an effective approach for employing GPT-4V to score drawn models. Even though there is space for GPT-4V to improve scoring accuracy, some mis-assigned scores seemed interpretable to science content experts. The results of this study show that utilizing GPT-4V for automatic scoring of student-drawn models in science education is promising, but there remains a challenging gap to further improve scoring accuracy.
}




\maketitle
\newpage
\section{Introduction}\label{introduction}

   Science centers "modeling the world," and thus engaging students in scientific modeling practices is critical to preparing students' science competence \citep{nrc2012framework, zhai2022assessing}. During modeling practices, students can draw models and visualize their ideas and explanations of phenomena, providing an avenue for students with diverse language proficiency to learn science in an equitable manner. These models are particularly valuable for delving into students' thinking, given the enriched information encapsulated in these models. However, teachers find them time- and effort-consuming to score, which presents challenges to introducing modeling practices in classrooms. Therefore, researchers have leveraged supervised machine learning to develop deep learning algorithmic models to score student responses \citep{zhai2022applying, lee2023automated}.
   
   %Constructed response (or free-response) items typically require students to draw a scientific model that connotes their thoughts on the given natural phenomena and then explain what the model represents in written responses. Therefore, students' responses to constructed items are multimodal in their nature, which conveys much more information on students' understanding of the content being taught, compared to multiple-choice or written response items \citep{Nyachwaya2011}. However, it has been frequently pointed out that assessing student-drawn models is time-consuming and labor-intensive, 
   
   Using this approach, human experts have to assign scores to student-drawn models using well-developed scoring rubrics. Usually, more than two human experts are needed, and their consensus and interrater reliability are considered essential to reduce rater errors and potential bias. \cite{hogan2007recommendations} suggest additional considerations such as scoring one item at a time, using a scoring rubric or ideal answer, and reading samples before scoring others. The human-scored data would be used to train machines to develop scoring models. Though researchers reported satisfied scoring accuracy, the entire algorithmic model development is usually costly and time-consuming. It is thus urgent to develop approaches to reduce the efforts of developing scoring models.

With the recent development of GPT-4V, \cite{openai2023gpt4} provides opportunities to employ prompt engineering to reduce machine training time and costs. GPT-4V takes an image and a natural-language prompt about the image as input and can produce a natural-language answer as the output \citep{Antol_2015_ICCV}. The capability of providing explainable answers in natural language to users' questions about the image is considered a milestone development for visual question answering \citep{joshi2021review}. This technology is deemed useful to science education researchers and practitioners, where constructed-response items that assess student-drawn models are frequently used. However, until now, there has been scarce research on image processing or visual question answering for automatic scoring in the science education field.

%In Semtermber 25th, 2023  officially announced that "ChatGPT can now see" on the ChatGPT service webpage with the GPT-4V model. As reviewed later, developers in Microsoft and scholars around the medical sciences are eagerly examining the applicability of GPT-4V in various problem-solving. However, there seems to be almost no research that applied GPT-4V for educational studies, particularly for automatic scoring of student-drawn image answers. Therefore, it is strongly requested to examine GPT-4V's capability in automatically scoring scientific model images.

Given this research gap, we developed a method NERIF (Notation-Enhanced Rubric Instruction for Few-shot Learning) employing \textit{instructional note} and rubrics to prompt GPT-4V to score students' drawn models for science phenomena. This study answers two research questions:
    %\begin{itemize}
    %\item 
    \textbf{1)} How accurate is GPT-4V in automatically scoring student-drawn models?
    %\item 
    \textbf{2)} How does GPT-4V automatically assign scores to student-drawn models?
    %\end{itemize}

\section {Automatic Scoring of Scientific Modeling Practices}\label{}
     Scientific modeling is a cornerstone of science education as it serves as a bridge between mental models and real-world phenomena. Engaging students in scientific modeling practices fosters a deeper understanding of the nature of science as an iterative and predictive process during which students use knowledge to explain phenomena \citep{hestenes2013modeling}. Models provide a framework for students to conceptualize phenomena, deploy scientific knowledge, and represent ideas and explanations \citep{zhai2022assessing}. This active involvement in the construction, testing, revision, and deploying of models also supports the development of critical thinking and problem-solving skills. Moreover, modeling equips students with the ability to communicate and justify their reasoning, reflecting the collaborative and communicative nature of scientific inquiry \citep{clement2008creative}. By embodying the practices of scientists, students gain a more authentic understanding of the nature of science, thus enhancing their ability to apply scientific knowledge to novel situations, a skill increasingly critical in our rapidly evolving world.
    
    Despite the importance of modeling practices in science learning \citep{nrc2012framework}, drawn models are rarely employed in science classrooms for assessment practices, partially because drawn models are challenging to score, and thus students rarely receive timely feedback for their specific models. One solution to this problem is employing automated scoring technologies to grade students' scientific modeling. Recent studies have leveraged machine learning (ML) techniques for the automatic evaluation of student-generated models. Notably, \cite{smith2018multimodal} focused on scoring models constructed by middle schoolers concerning magnetic concepts. Their methodology involved the use of pre-determined elements within a structured digital environment, allowing students to manipulate these elements spatially. They utilized a topology-centric scoring method that recognized spatial relationships—proximity, distance, and containment—between elements. Additionally, specific rules constrained possible relations to refine the scoring accuracy. However, \cite{smith2018multimodal} technique has limitations, particularly in handling the unpredictability inherent in free-form drawing tasks.

    Addressing the challenges presented by unstructured drawings, \cite{von2023scoring} implemented advanced deep learning strategies to assess an exercise from the Trends in International Mathematics and Science Study (TIMSS). This task provided a gridded canvas for students to depict their conceptual understanding through drawing, with the grid serving as a constraint to reduce variation in the drawings. \cite{von2023scoring} adopted convolutional neural networks (CNNs) and feed-forward neural networks (FFNs), two distinct artificial neural network architectures with different data processing and learning capabilities. Their findings indicated the superior performance of CNNs over FFNs in scoring accuracy, and notably, CNNs demonstrated an ability to outperform human scoring in certain instances where humans had misjudged responses, highlighting the potential of ML in educational assessments. Given the progress, the models in \cite{von2023scoring} are not free-drawing, which limits the usability in science classrooms. 
    
    In their study, \cite{zhai2022applying} developed six assessment free-drawn modeling tasks. These tasks are embedded in computer-simulated environments and provide a drawing pad for students to represent their models. The tasks "ask students to draw a model to make sense of the phenomena using online tools," targeting an NGSS performance expectation for middle school students: "\textit{MS-PS1-4. Develop a model that predicts and describes changes in particle motion, temperature, and state of a pure substance when thermal energy is added or removed}" (p. 1774). The analytic scoring rubric provides principles to categorize students' drawn models into 'Beginning,' 'Developing,' or 'Proficient' levels. They employed ResNet-50 V2 CNN to develop scoring models and tested the scoring models with more than 250 new student-drawn models for each of the six tasks. The research reported machine-human scoring agreement of accuracy in .79-.89 and Cohen's Kappa in .64-.82.

   To be noted, the above research leveraged computers and asked students to draw models on computers. In real classroom settings, teachers often ask students to draw models on papers, which adds more degree of freedom and can be more challenging for automatic scoring. \cite{lee2023automated} developed a model that automatically assesses elementary, middle, and high school students' responses to the two items adopted from Test About Particles in a Gas \citep{novick1981pupils}. The authors embedded students' hand drawings using Inception-v3 pre-trained model \citep{szegedy2016rethinking}, and tried various machine learning algorithms such as k-nearest neighbor, decision tree, random forest, support vector machine, neural network, logistic regression as the final classifier layer. As results with 206 test cases, they reported that their model performance reached a high machine-human agreement (kappa = 0.732–0.926, accuracy = 0.820–0.942, precision = 0.817–0.941, recall = 0.820–0.942, F1 = 0.818–0.941, and area under the curve = 0.906–0.990).

   Also, \cite{wang2024modeling} employed 2D convolutional neural networks to automate the assessment of high school students' hand-drawn models on the topic of optics. They analyzed 758 student-created models explaining the refraction phenomenon. Employing a sequential ML model composed of four convolutional layers, they attained a commendable average accuracy rate of 89\% (SD=9\%). Further, nested cross-validation yielded an average testing accuracy of 62\% (SD=5\%), with notable accuracy discrepancies observed across groups with varying modeling proficiency levels. Intriguingly, models from students with lower performance proved more difficult for the ML algorithms to score accurately. They conducted a comparative analysis of the models, distinguishing between those consistently scored correctly and those frequently misjudged by the machine. Their investigation revealed that certain characteristics inherent in the students' drawn models were influential in the machine's scoring precision. 

    These previous studies exemplify that there it is possible to automatically assess students' drawing models on natural phenomena by applying prominent ML techniques with computer vision. However, techniques used in their studies, such as ResNet 50 V2 or Inception-V3 pre-trained model, can be technical barriers to researchers with less machine learning expertise. Therefore, it is necessary to explore ways to broaden the usability of computer vision techniques to the larger group of the education community, and a visual language model such as GPT-4V is a potential candidate for that initiative. Moreover, the supervised approach for scoring model development is time and cost-consuming and needs new methods to overcome these challenges.

\section {GPT-4V for Image Processing}\label{}

ChatGPT, a state-of-the-art large language model, has had tremendous impacts on and changed education. Among the many applications in education \citep{Lo2023, Grassini2023}, ChatGPT and GPT API have shown significant advantages in automatic scoring to facilitate timely feedback and personalized learning \citep{zhai2022chatgpt,zhai2023chatgpt, zhai2023chatgpt1}. For example, \cite{latif2023fine} have leveraged the powerful natural language processing, understanding, and generating ability of the GPT family and fine-tuned ChatGPT-3.5 turbo to accurately score student written explanations for science phenomena, which shows 9.1\% average increase of scoring accuracy compared to BERT. In addition, GPT's powerful generative ability could help solve challenging automatic scoring problems such as data imbalance. Unbalanced training data can introduce scoring uncertainty and result in biased outcomes. \cite{fang2023using} employed GPT-4 to augment unbalanced training data and found the GPT-generated responses yield identical outcomes compared to authentic student-written responses. Using this data augmentation method, they reported an increase of 3.5\% for accuracy, 30.6\% for precision, 21.1\% for recall, and 24.2\% for F1 score. Also, \cite{Kieser2023} showed that ChatGPT can emulate university students' written answers on Force Concept Inventory, scoring almost equivalent to those.

One notable development of OpenAI is the release of GPT-4V, which integrates an image processing module to GPT-4, enabling visual question answering (i.e., receiving textual prompt and image input and answering the user's question about the image in natural language). Visual ChatGPT, the predecessor of GPT-4V, was developed by Microsoft developers \citep{wu2023visual}, which incorporated various Visual Foundation Models such as Visual Transformers or Stable Diffusion to ChatGPT. They reported that Visual ChatGPT could generate images, extract features from the input image, change an object in the image with re-drawing, etc., following the user's natural language-based queries. After the release of the GPT-4V model, researchers from Microsoft \citep{yang2023dawn} made an initial but comprehensive report on the image processing ability of GPT-4V. \cite{yang2023dawn} introduces GPT-4V's working modes and prompting techniques, GPT-4V's vision-language capability, temporal understanding, intelligence and emotional quotient tests, etc.  \cite{wu2023early} further presented varying abilities of GPT-4V in domains such as visual understanding, language understanding, and visual puzzle solving. However, their focus was not on the image classification performance of models.

Scholars have reported the application of GPT-4V in various problem domains, although there is no open GPT-4V API as of now (November 5th, 2023). Most of these early reports were made by medical scholars in medical subdomains. For example, \cite{wu2023gpt4v} evaluated GPT-4V's diagnosing ability in human body systems such as the central nervous system, cardiac, obstetrics, etc. They descriptively suggested that GPT-4V demonstrated proficiency in distinguishing medical image modalities and anatomy and showed difficulties in diagnosing symptoms. \cite{wang2023bioinformatics} reported that GPT-4V can read Bioinformatics illustrations such as sequencing data analysis, multimodal network-based drug repositioning, and tumor clonal evolution - however, it showed weakness in quantitative counting of visual elements. 

\cite{chen2023gpt4}'s study is noteworthy in that they provided quantitative results of GPT-4V's classification of images. They used the Kaggle COVID-19 lung X-ray dataset for a binary classification problem (COVID-19 vs normal cases). According to \cite{chen2023gpt4}, GPT-4V (accuracy = .72-.83) outperformed ResNet (accuracy = .74) and VGG models (accuracy = .80) in 6-shot learning situation. However, both models trained with the full dataset performed better than GPT-4V. \cite{Yang2023performance} also demonstrated the performance of multimodal GPT-4V on USMLE question bank, which marked 86.2\% accuracy.

In other fields, \cite{chen2023can} tested whether GPT-4V completely masks personal information and found that it starts to reveal the correct location of a building in the given image. \cite{chen2023towards} examined GPT-4V's decision-making ability in automatic driving, housing robot, and open-world game domains.  Also, \cite{zhou2023vision} showed the possibility that GPT-4V could be used for evaluating traffic anomaly scenes. Meanwhile, \cite{shi2023exploring} examined the optical character recognition performance of GPT-4V and confirmed that it recognized and understood Latin-alphabet content well but was struggled in recognizing other character systems-written content.
    
To sum up, there have been very few studies that tested the classification performance of GPT-4V. Further, no study has explored the possibility of applying GPT-4V for educational studies, particularly for the automatic scoring of multinomial items.

\section{Methods}\label{}

    \subsection{Data Set}

This study secondarily analyzed student-drawn scientific models from a dataset adapted from a parent study by \cite{zhai2022applying}. The items developed by the NGSA team \citep{Harris2024Creating} target the  \textit{NGSS} \citep{ngss2013next} performance expectations, to implement a 3-Dimensional assessment that incorporates disciplinary core ideas, cross-cutting concepts, and science and engineering practices.
    
    We conducted an experiment with the six items. Per each item, we randomly sampled 9 example human evaluation cases, nine validation cases, and 150 test cases. The validation and test datasets were perfectly balanced throughout the categories (1/3 for 'Proficient,' 1/3 for 'Developing,' and 1/3 for 'Beginning'). 
    
    \subsection{Experimental Design}

\begin{figure}
        \centering
        \includegraphics[width=1\linewidth]{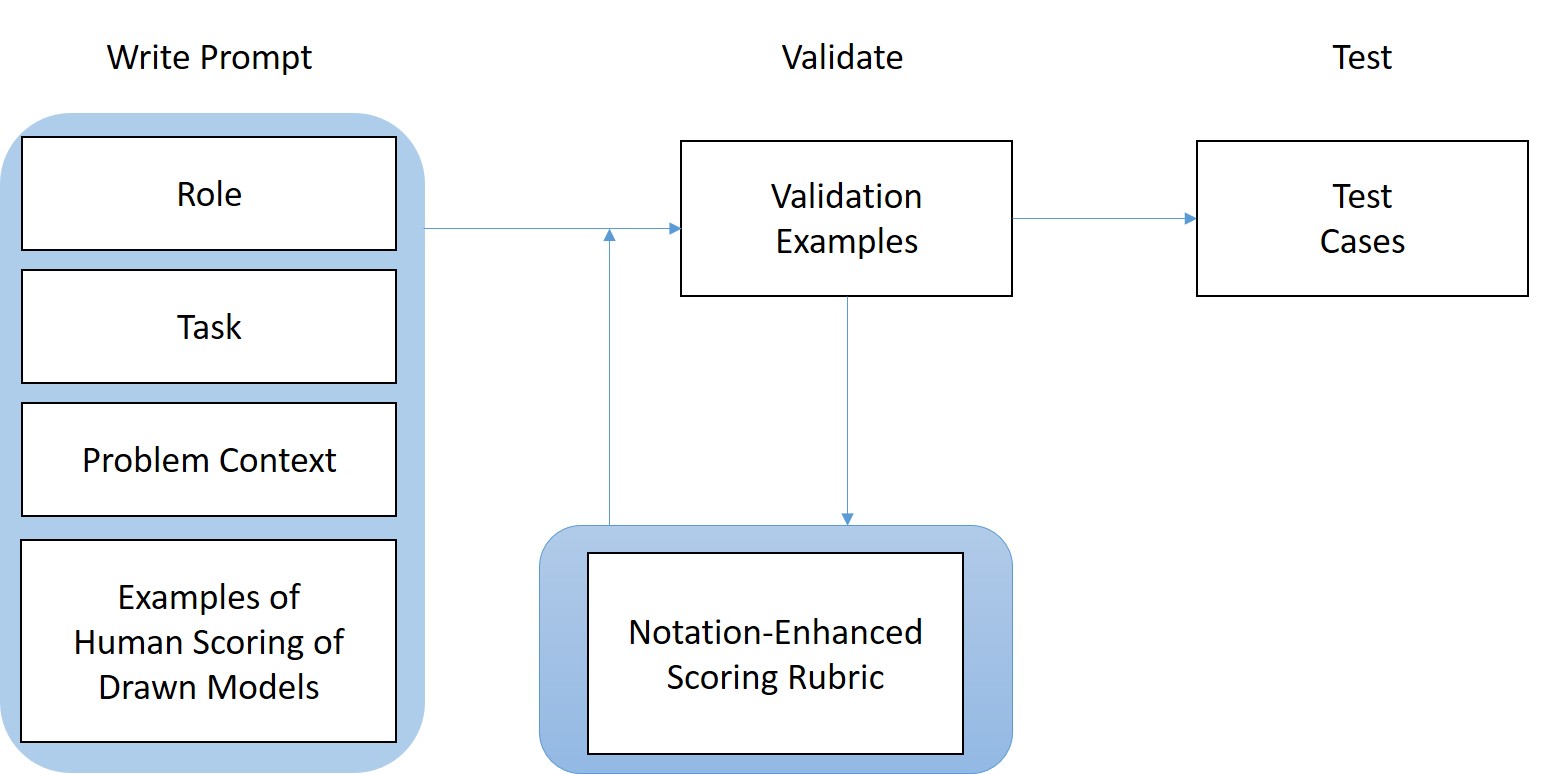}
        \caption{The Process of Notation-Enhanced Rubric Instruction for Few-shot Learning}
        \label{fig:NERIF_Process}
    \end{figure}

    The goal of the experiment was to develop a method, NERIF (Figure \ref{fig:NERIF_Process}), that can help users utilize GPT-4V for automatic classification of image data and test it with student-drawn scientific models.

\textit{    Write Prompt}. To achieve this goal, we used a few-shot learning approach \citep{wang2020generalizing, wu2023early} with 9 example evaluations to instruct GPT-4V to correctly categorize student-drawn visual answers according to the scoring rubric. The task GPT-4V should solve is a multinomial classification in step with the given data.

    %\begin{itemize}
    %\item 
    \textit{Validation}. We confirmed that our prompt and input image data could instruct ChatGPT to read the given images and categorize student-drawn models with the validation dataset (\textit{N} = 54). The validation step also served for heuristic prompt engineering of notation-enhanced scoring rubric.

   %\item 
   \textit{Test}. After validation, we repeatedly ran a GPT-4V session to automatically score student-drawn images in the test dataset (\textit{N} = 900). By setting the temperature to 0.0 and top\_p to 0.01, we intended to receive the most reliable results from ChatGPT (greedy decoding).
   %\end{itemize}

    \subsection{NERIF: Notation-Enhanced Rubric Instruction for Few-shot Learning}

\begin{figure}[htbp]
    \vspace{-0.2cm}
    \centering
    \includegraphics[width=0.9\linewidth]{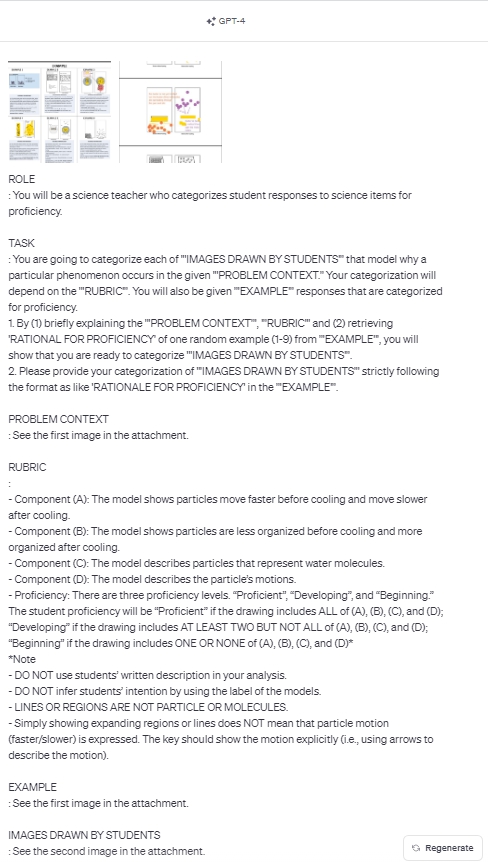}
    \vspace{-0.2cm}
    \caption{Example Prompt (Task M3-1)}
    \label{Fig:examplePrompt}
    \end{figure}

\begin{figure}[htbp]
    \vspace{-0.2cm}
    \centering
    \includegraphics[width=0.9\linewidth]{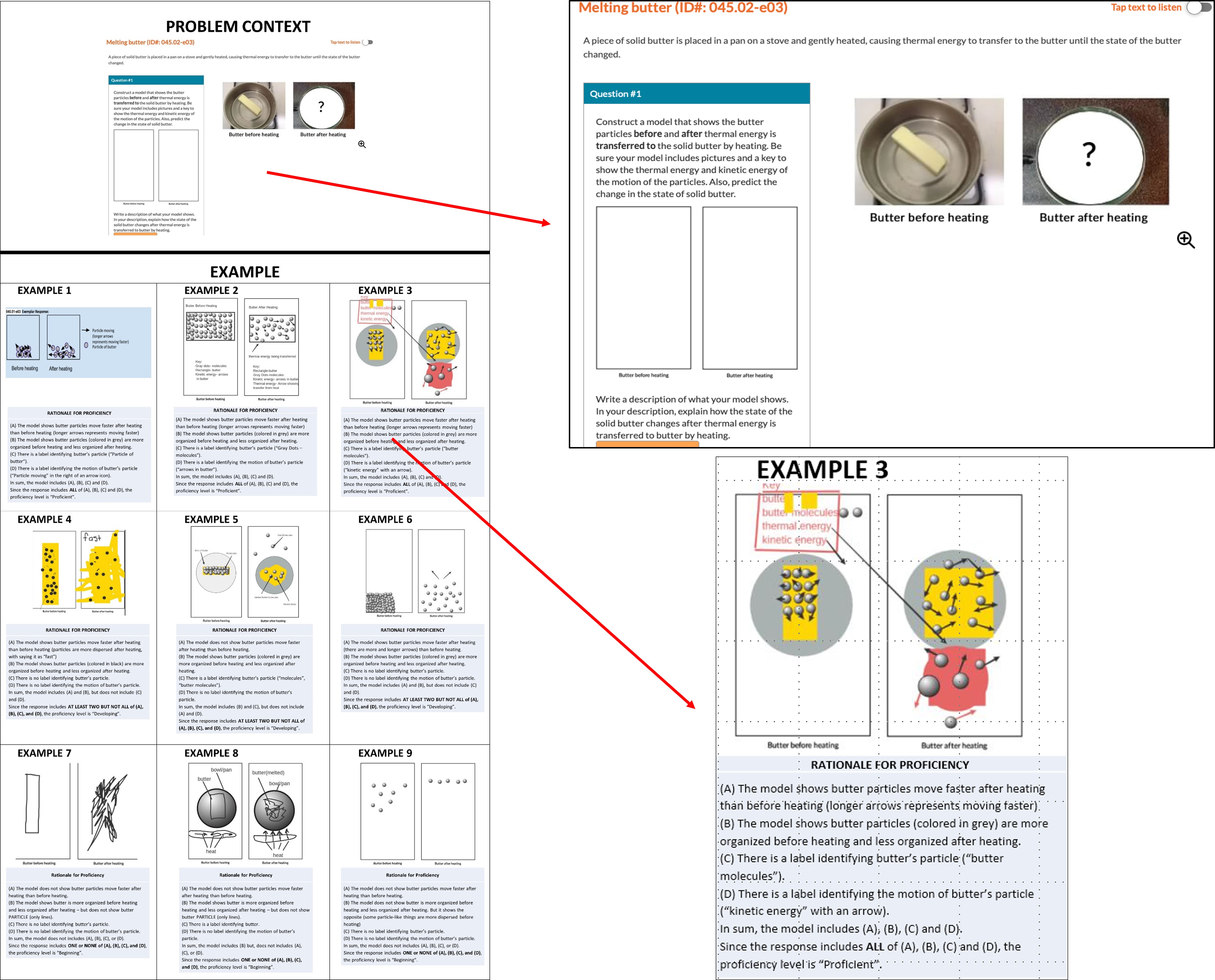}
    \vspace{-0.2cm}
    \caption{Example Input Image 1 (Task M3-1) - Problem Context and Scoring Examples}
    \label{Fig:exampleImage1}
    \end{figure}

    We gave ChatGPT two images with a prompt as input for the automatic scoring of student-drawn models. The first images includes a problem statement, and 9 example human coders' assessments on student-drawn models for few-shot learning. The second image included three student-drawn models that ChatGPT was instructed to assess.
    Our prompt for single-turn conversation consisted of 7 components. An example of our prompt and input image for processing students' scientific modeling is presented in Figures \ref{Fig:examplePrompt}-\ref{Fig:exampleImage1}, respectively.

    %\begin{itemize}
    
   % \item 
    \textbf{(1)} \textit{Role} designates in what position ChatGPT should answer to the query. \textit{Role} was given as "You will be a science teacher who categorizes student responses to science items for proficiency." 
    
    %\item 
    \textbf{(2)} \textit{Task} explains what ChatGPT is requested to do. ChatGPT's task is to categorize \textit{models drawn by students} that model why a particular phenomenon occurs in the given \textit{problem context}. Also, ChatGPT's categorization must depend on the \textit{rubrics}. ChatGPT should learn how to categorize the models and provide the 'rationale for proficiency' from the human coders' demonstration, which is given in \textit{example}. After instructing that:
        %\begin{itemize}
        %\item 
        %\textit{a)} 
        \textit{Task} requires ChatGPT to retrieve \textit{problem context}, \textit{rubric}, and 'rationale for proficiency' of one random example from \textit{example}. Note that \textit{problem context} and \textit{example} are given in the first image attached.
        %\item 
        %\textit{b)} 
        \textit{Task} then requests ChatGPT to categorize \textit{models drawn by students}, with its 'rational for profeciency' as like in \textit{example}.
        %\end{itemize}
    %\item 
   
    \textbf{(3)} \textit{Problem context} is given in the first image attached, and ChatGPT has to retrieve it. For example, Task M3-1 is contextualized in a scenario which students are heating a solid butter until the state of the butter changes. The item requires students to construct a model that shows before and after thermal energy is transferred to the solid butter by heating.
    
    %\item 
    \textbf{(4)} \textit{Notation-Enhanced Scoring Rubrics} includes three components to guide GPT-4V for automatic scoring:
        %\begin{itemize}
        %\item 
        
        \textit{Human experts identifying scoring aspects}. We asked human subject matter experts to specify the aspects that should be considered to assess student-drawn models. The scoring rubric for items used in this study considers up to 2-4 components. For example, Task M3-1 considers four components for scoring (see Figures \ref{Fig:examplePrompt}-\ref{Fig:exampleImage1}).
        
        %\item 
        \textit{GPT-4 defining scoring rules aligned with proficiency levels}. \textit{Proficiency} defines the rule to categorize student-drawn models, synthesizing the aspects the drawing includes. The proficiency level is trinomial categories of 'Proficient', 'Developing', and 'Beginning'. We first ask GPT-4 to identify which aspect(s) are included in the scoring rule for each specific proficiency level, to help ChatGPT categorize the test case according to the \textit{Proficiency} rules.
        For example, in Task M3-1, the student's answer is considered 'Proficient' only when \textit{all} the four components are included, 'Developing' when \textit{at least two but not all} of the four are included, and 'Beginning' when \textit{one or none} of those are included.
        
        %\item 
        \textit{Instructional Notes} provides ChatGPT with hints to decide whether an image includes each component or not. Notes were heuristically found and written by researchers during the validation step.
        %\end{itemize}

%\item 
    \textbf{(5)} \textit{Example} are given for few-shot learning. Each example consists of one student-drawn model and one human coder's labeling of that model, with 'Rationale of Proficiency' level labeling according to the scoring rubric. It helps GPT-4V to summarize what it has decided on each component and deduce appropriate category from the \textit{Rubrics}. We designed this feature inspired by the Chain-of-Thought technique \citep{wei2022chain, wu2023visual}. A total of nine examples were included in the below of the first attached image.
    
    %\item 
    \textbf{(6)} \textit{Models drawn by students} are test cases. Total of three student drawn scientific model were included in the second attached image.
    
    %\item 
    \textbf{(7)} \textit{Temperature/top\_p} are hyper-parameters for ChatGPT's response.
    %\end{itemize}

    \subsection{Data Analysis}

    We repeatedly ran GPT-4V sessions to analyze the test cases. GPT-4V assessed three student-drawn models per query we sent, which included the two images and prompt. We opened a new session for every three students' drawn models, which ChatGPT assesses in a turn of conversation, lest GPT-4V's memorization of conversation affect the assessment of later test cases. After collecting GPT-4V's assessment of images drawn by students, accuracy, precision, recall, F1, and Fleiss' Kappa were calculated by comparing GPT-4V scores with the human scores. Further, the two researchers of this study, who are experts in science education and automatic assessment, inductively identified the characteristics of GPT-4V's behaviors to uncover the scoring process during the experiment.

% \begin{figure*}[t]
%     \centering
% \includegraphics[width=0.50\linewidth]{AGI_core.pdf}
% \caption{A microscopic view of AGI Core}
%     \label{fig:agi_core}
%     \vspace{-2mm}
% \end{figure*}

\section {Findings} \label{Findings}
In this section, we first present scoring accuracy, including the accuracy parameters for both validation and testing processes. Then we report the GPT-4V scoring processes with examples, and uncover notable behavior patterns of GPT-4V in scoring the models.

    \subsection {GPT-4V Scoring Accuracy on Student Drawn Models}
    
    \subsubsection{Validation Scoring Accuracy}

    The validation process was used to help researchers develop and revise the prompts, which iteratively changed as we improve the prompts. Here we reported the final validation accuracy for our prompts for the six items (seeTable \ref{tab:Val_Accuracy}). The accuracy for the 'Beginning' examples was .78, for 'Developing' was .67, and for 'Proficient' was .56. On average, our prompts showed validation accuracy of .67 in assessing student-drawn models.

%%%%%%%%%%%%%%%%%%%%%%%%%%%%%%%%%%%%%%%%%%%%%%%%%%%%%%%%%%%%%%%%%%%%%%%%%%%%%%%%%%%%%%%%%%%%%%%
\begin{table}[ht]
\centering
\caption{Validation Accuracy of GPT-4V for drawing assessment}
\begin{tabular}{lcccc}
\hline
\textbf{Item} & \textbf{\begin{tabular}[c]{@{}c@{}}Overall      (\textit{N} = 9)\end{tabular}}& \textbf{\begin{tabular}[c]{@{}c@{}}Beginning      (\textit{n} = 3)\end{tabular}} & \textbf{\begin{tabular}[c]{@{}c@{}}Developing      (\textit{n} = 3)\end{tabular}} & \textbf{\begin{tabular}[c]{@{}c@{}}Proficient      (\textit{n} = 3)\end{tabular}} \\ \hline
R1-1& 0.78                                                                                  & 1.00                                                                                    & 1.00                                                                                     & 0.33                                                                                     \\
J2-1& 0.67                                                                                  & 1.00                                                                                    & 0.67                                                                                     & 0.33                                                                                     \\
M3-1& 0.56                                                                                  & 0.67                                                                                    & 0.33                                                                                     & 0.67                                                                                     \\
H4-1& 0.89                                                                                  & 0.67                                                                                    & 1.00                                                                                     & 1.00                                                                                     \\
H5-1& 0.67                                                                                  & 0.67                                                                                    & 0.67                                                                                     & 0.67                                                                                     \\
J6-1& 0.44                                                                                  & 0.67                                                                                    & 0.33                                                                                     & 0.33                                                                                     \\ \hline
Mean       & 0.67                                                                                  & 0.78                                                                                    & 0.67                                                                                     & 0.56                                                                                     \\ \hline
\end{tabular}
\label{tab:Val_Accuracy}
\end{table}
%%%%%%%%%%%%%%%%%%%%%%%%%%%%%%%%%%%%%%%%%%%%%%%%%%%%%%%%%%%%%%%%%%%%%%%%%%%%%%%%%%%%%%%%%%%%%%%

    \subsubsection{Test Scoring Accuracy}

    To examine scoring accuracy, we tested the prompts with new samples that GPT-4V did not see during the prompt development phase. Table \ref{tab:Test_Accuracy} shows the test accuracy for the 6 items. On average for the six tasks, GPT-4V yielded a test scoring accuracy of .51. (SD = .037), with the average value of Precision =.58 (SD = .04), Recall = .51 (SD = .037), and F1 = .49 (SD = .047). Fleiss' Kappa (quadratic weighted) ranged from .32 to.51, which is considered 'Fair' to 'Moderate' accuracy \citep{LandisKoch1977}.

%%%%%%%%%%%%%%%%%%%%%%%%%%%%%%%%%%%%%%%%%%%%%%%%%%%%%%%%%%%%%%%%%%%%%%%%%%%%%%%%%%%%%%%%%%%%%%%

\begin{table}[]
\centering
\caption{Testing Scoring Accuracy of GPT-4V for drawing assessment}

\begin{tabular}{p{1.5cm}p{1cm}p{1cm}p{1cm}p{1cm}p{1cm}p{1cm}p{0.7cm}p{0.9cm}}

\toprule
%\hline
\textbf{Item} & \textbf{Accuracy} & \textbf{Acc\_Beg} & \textbf{Acc\_Dev} & \textbf{Acc\_Prof} & \textbf{Precision} & \textbf{Recall} & \textbf{F1} & \textbf{Kappa} \\ %\hline
\midrule
R1-1& 0.50              & 0.50              & 0.66              & 0.34               & 0.56               & 0.50            & 0.50        & 0.44           \\
J2-1& 0.45              & 0.68              & 0.56              & 0.12               & 0.62               & 0.45            & 0.41        & 0.32           \\
M3-1& 0.53              & 0.82              & 0.40              & 0.36               & 0.53               & 0.53            & 0.51        & 0.51           \\
H4-1& 0.57              & 0.64              & 0.68              & 0.38               & 0.61               & 0.57            & 0.56        & 0.51           \\
H5-1& 0.47              & 0.62              & 0.58              & 0.22               & 0.53               & 0.47            & 0.46        & 0.43           \\
J6-1& 0.53              & 0.62              & 0.84              & 0.12               & 0.62               & 0.53            & 0.48        & 0.38           \\ %\hline
Mean       & 0.51              & 0.65              & 0.62              & 0.26               & 0.58               & 0.51            & 0.49        & 0.43           \\ %\hline

\bottomrule
\end{tabular}

\label{tab:Test_Accuracy}
\end{table}
%%%%%%%%%%%%%%%%%%%%%%%%%%%%%%%%%%%%%%%%%%%%%%%%%%%%%%%%%%%%%%%%%%%%%%%%%%%%%%%%%%%%%%%%%%%%%%%

    We also found that the scoring accuracy vary by scoring category. Specifically, the accuracy for the 'Beginning' cases was .64, for 'Developing' cases was .61, and for 'Proficient' cases was .26 (see Table \ref{tab:Test_Accuracy}). To understand the variations, we delve into the confusion matrices of two example Tasks J2-1 and J6-2 (see Table \ref{tab:Confusion_Matrix}). For both items, GPT-4V predicted most of the true 'Beginning' and 'Developing' cases correctly. However, GPT-4V predicted the majority of 'Proficient' cases to be 'Developing'.

    \begin{table}
    \centering
    \caption{Confusion Matrix of Tasks J2-1 and J6-1}
    \begin{tabular}{lcccccc}
    \hline
    \multirow{2}{*}{\textbf{True Label}} & \multicolumn{3}{c}{\textbf{Task J2-1}} & \multicolumn{3}{c}{\textbf{Task J6-1}} \\
                                         & Beginning  & Developing & Proficient & Beginning  & Developing & Proficient \\ \hline 
    Beginning                            & 34         & 16         & 0          & 31         & 19         & 0          \\
    Developing                           & 22         & 28         & 0          & 6          & 42         & 2          \\ \
    Proficient                           & 12         & 32         & 6          & 8          & 36         & 6          \\ \hline
    \end{tabular}
    \label{tab:Confusion_Matrix}
    \end{table}

    \subsection{Unpacking How GPT-4V Score Student Models} \label{CaseStudy}

\begin{figure}
    \centering
    \includegraphics[width=0.725
    \linewidth]{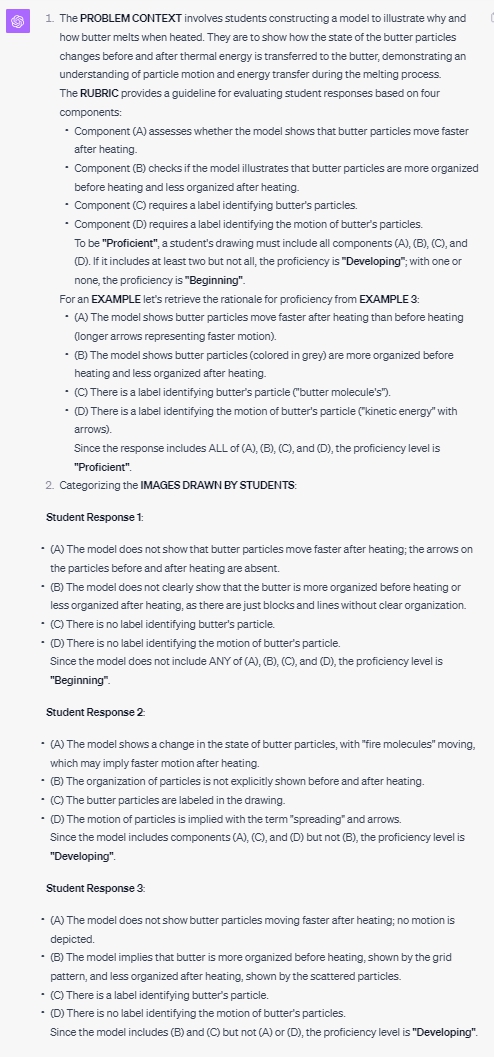}
    \caption{Example Response of GPT-4V (Task M3-1)}
    \label{fig:exampleResponse}
\end{figure}

\begin{figure}[htbp]
    \vspace{-0.2cm}
    \centering
    \includegraphics[width=1.1\linewidth]{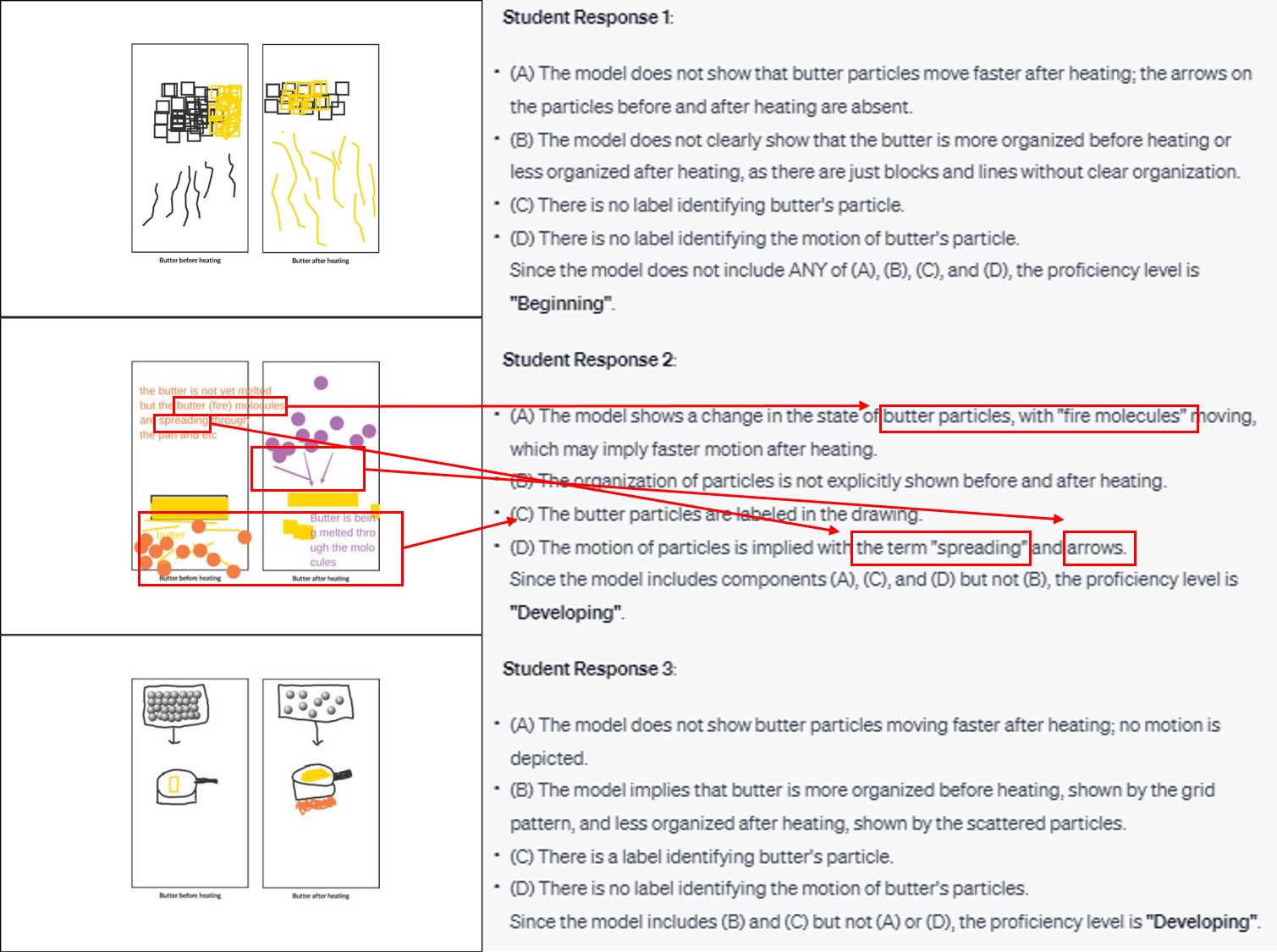}
    \vspace{-0.2cm}
    \caption{Example Prediction of GPT-4V (Task M3-1) on Three 'Developing' Test Cases}
    \label{Fig:examplePrediction}
    \end{figure}

   To uncover GPT-4V's scoring process, we present an example in Figure \ref{fig:exampleResponse}, which shows examples of GPT-4V scoring on three students' drawn models. The input given to ChatGPT was the prompt presented in Figure \ref{Fig:examplePrompt}, problem context and scoring examples presented in Figure \ref{Fig:exampleImage1}, and three images of student-drawn scientific models, which were concatenated into one image file (Figure \ref{Fig:examplePrediction}).

        In the example response, ChatGPT performs the \textit{task} defined in the prompt (Figure\ref{Fig:examplePrompt}). It first briefly explains the \textit{problem context} and \textit{rubric}, and then retrieves a random \textit{example} of human coders' evaluation of the student's scientific model.
        
    ChatGPT's categorization of the three student-drawn models follows, which is structured as the 'rationale for proficiency' in the \textit{example}s. It determines whether each component defined in the rubric is included in the test cases, summarizes its evaluation, and deduces the final categorization (one of 'Beginning,' 'Developing,' and 'Proficient'). This again shows that GPT-4V can retrieve images given as input and separately process them according to the user's query.

    The example prediction of GPT-4V on the three 'Developing' test cases is presented in Figure \ref{Fig:examplePrediction}. It is observed that GPT-4V correctly predicted two of the three as 'Developing.' and one as 'Beginning.' Further contemplation of GPT-4V's analysis of student-drawn models is presented in the section \ref{CaseStudy}. Below we present sevearl features of GPT-4V identified in our qualitative analyses.

\textbf{\textit{GPT-4V can recognize and retrieve information from the questions presented in images}}. We found that GPT-4V can successfully access the input images and process information encapsulated (see examples in Figures \ref{fig:exampleResponse} and \ref{Fig:examplePrediction}). Both the \textit{problem context} and \textit{example} are provided in the format of the image and GPT-4V successfully retrieved the questions and answers in the input image. Particularly, GPT-4V stringently followed our instruction to retrieve one random example from the input image (Figure \ref{Fig:exampleImage1}) and treated it as evidence of its processing of the given image (Figure \ref{fig:exampleResponse}). This precise retrieval of information strengthens the GPT-4V's possible use for automatic scoring.

\textbf{\textit{GPT-4V can catch the characteristics of student-drawn models}}. Figure \ref{Fig:examplePrediction} exemplifies that GPT-4V is able to capture characteristics of student-drawn scientific models. It read the printed student descriptions of their models, such as "butter particles," "fire molecules," and "spreading." Also, it rightly pointed out there is an arrow in the student-drawn image, which signifies the motion of particles.

\begin{figure}
    \centering
    \includegraphics[width=1\linewidth]{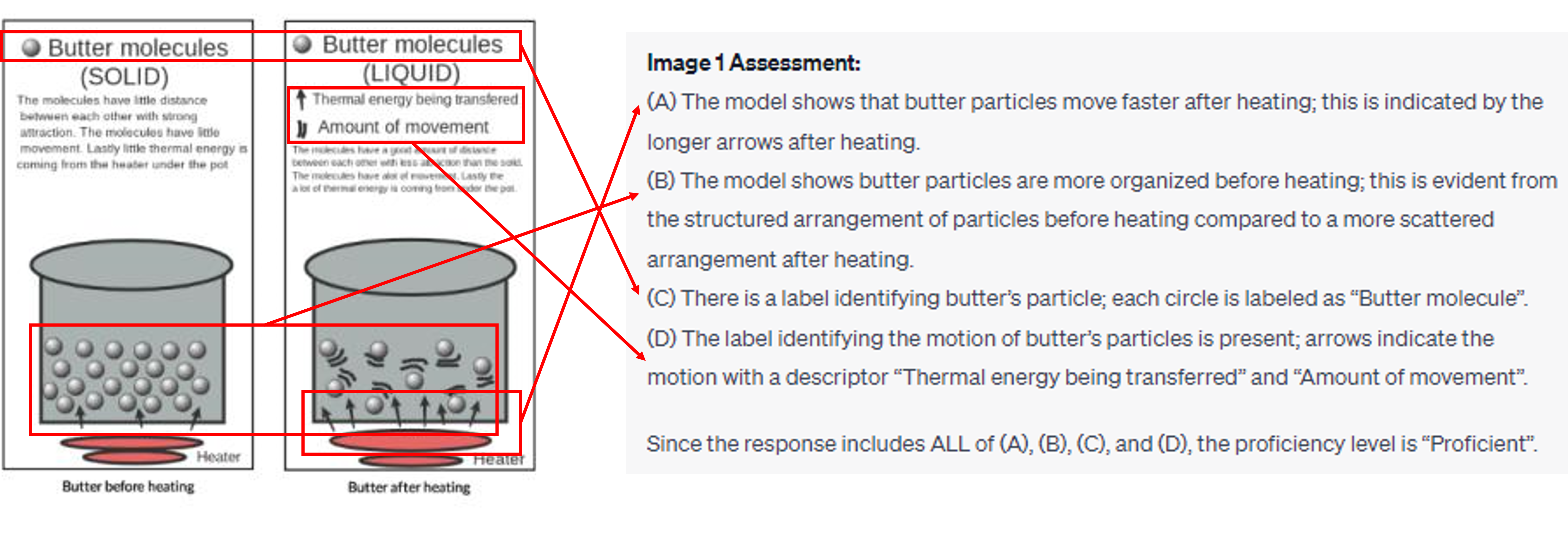}
    \caption{Example Prediction of GPT-4V (Task M3-1) on One 'Proficient' Test Case}
    \label{fig:exampleProficientInteresting}
\end{figure}

Another example of GPT-4V's understanding of students' models is presented in Figure \ref{fig:exampleProficientInteresting}. GPT-4V takes the "longer arrows after heating" in the image as evidence for component (A). GPT4-V interprets the image as "the structured arrangement of particles before heating compared to a more scattered arrangement after heating," which denotes component (B). GPT-4V reads "Butter molecules" that label (gray) circle in the image to decide whether it includes a component (C). And GPT-4V read "arrows indicate the motion with a descriptor "Thermal energy being transferred" and "Amount of movement." Consequently, GPT-4V correctly predicted the image as 'Proficient.'

These examples show especially that GPT-4V not only extracts features from the image but also represents it in natural language, which the human user can understand why it made such a decision.

\textbf{\textit{GPT-4V can assess student-drawn models according to the rubric}}. Figures \ref{fig:exampleResponse}  and \ref{Fig:examplePrediction} also show that GPT-4V can assess student's visual answers according to the given rubric. GPT-4V relates the features that are extracted from the image to the appropriate components given in the rubric. For example, in Figure \ref{Fig:examplePrediction}, GPT-4V identified the changes in the butter particle's state (Component (A)), labels for butter particle (Component (C)), and keys such as "spreading" or arrows that describe butter particles' motion (Component (D)). However, it could not identify the changes in the organization of butter particles before and after the heating of butter (Component (B)). GPT-4V summarized that "the model includes components (A), (C), and (D), but not (B), the proficiency level is "Developing."", which corresponds to the scoring rubric and the human-coded category. This is apparent evidence that GPT-4V assesses student-drawn scientific model images as the given aspects and synthesizes them according to the rule.

\begin{figure}
    \centering
    \includegraphics[width=0.7\linewidth]{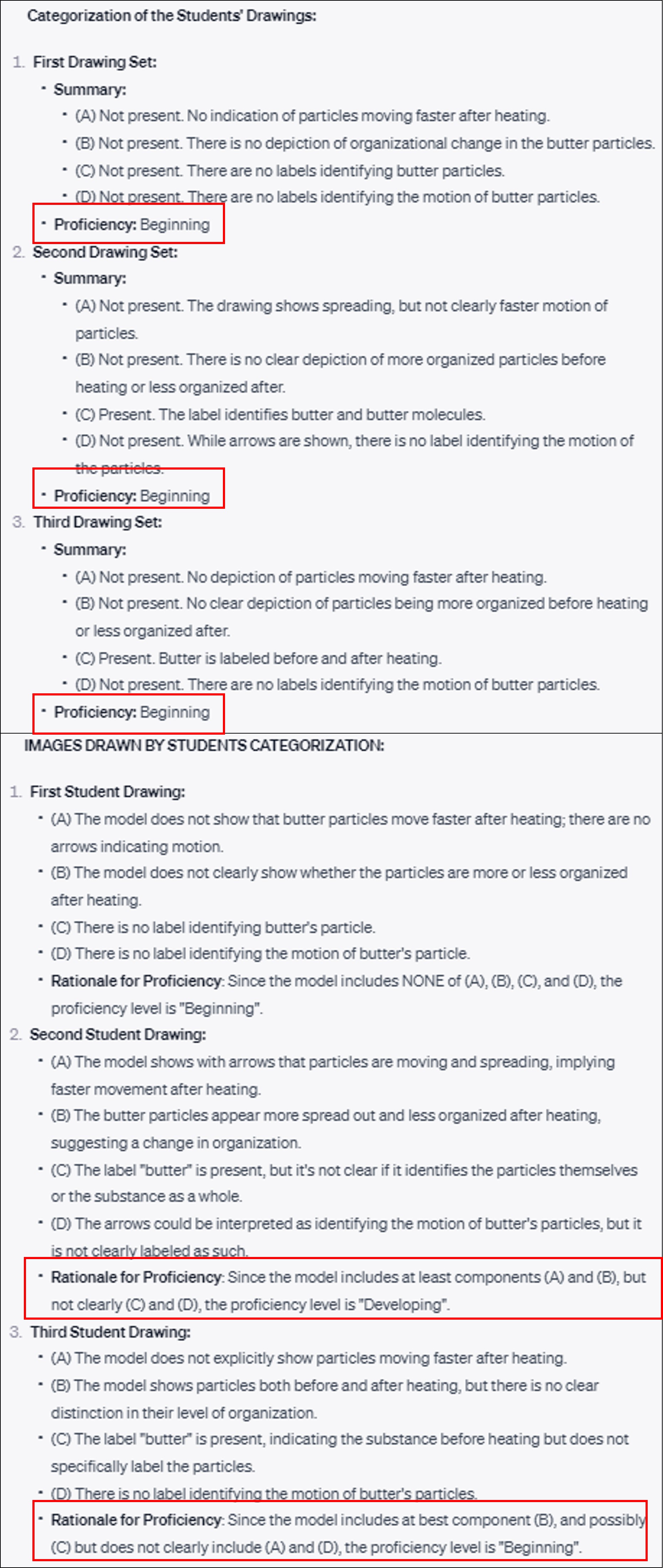}
    \caption{Example Prediction of GPT-4V (Task M3-1) on Three 'Developing' Test Cases, without \textit{notes} (top) or \textit{notes} (bottom)}
    \label{fig:No_Example_No_Notes}
\end{figure}

\textbf{\textit{Examples and notes can improve GPT-4V performance on scoring}}. What we have found during this study is that GPT-4V can be instructed to increase the quality of its inference on student-drawn images by \textit{example} and \textit{notes}. Figure \ref{fig:No_Example_No_Notes} presents the GPT-4V's prediction on the three 'Developing' test cases, which are same as those of Figure \ref{Fig:examplePrediction}.

The top of Figure \ref{fig:No_Example_No_Notes} shows that when the \textit{example} is not provided, the response of GPT-4V becomes very short, like "not present" and "present," which does not provide much information on the student-drawn images. It predicts that all the cases belong to the 'Beginning' category, which dramatically decreases the test accuracy. However, when there was \textit{example}, the response of GPT-4V was more elaborated, and it correctly predicted 2/3 cases to be 'Developing' (Figure \ref{Fig:examplePrediction}). This is clear evidence that \textit{example} provided in the attached image work as few-shot learning examples, which instructs GPT-4V how to assess student-drawn images.

The bottom of Figure \ref{fig:No_Example_No_Notes} shows that when the \textit{notes} is not provided, GPT-4V's prediction on each component defined in the rubric could be changed, and thus also the final predicted label. For example, GPT-4V decided that there was a component (C) in the second and third drawings when there was \textit{notes} (Figure \ref{Fig:examplePrediction}). However, when there was no \textit{notes}, GPT-4V's decision changed, and it judged that the drawings do not include component (C)). Consequently, the predicted label of the third drawing became 'Beginning,' which belongs to 'Developing' according to human coders or when there is \textit{notes}. This is a concrete example of guiding GPT4-V to appropriately process and categorize images by natural language-based instruction.

\textbf{\textit{GPT-4V sometimes makes incorrect but interpretative inferences}}. Although Figure \ref{fig:exampleProficientInteresting} shows that GPT-4V can correctly predict a student-drawn image's label, it also includes some interesting points that allow us to figure out how it made such a decision. GPT-4V insisted that "longer arrows after heating," which is indeed in the image, denotes "that butter particles move faster after heating." However, the sign of the butter particle's movement in the particular image in Figure \ref{fig:exampleProficientInteresting} was double lines (=) rather than arrows (→), as indicated by the student's explanation. Further, GPT-4V identified the arrows and double lines as the same symbol, saying that "arrows indicate the motion with a descriptor," "Thermal energy being transferred," and "Amount of movement." However, this is not true when we carefully read the student's comment in the image. It seems that the student intended to signify the transfer of thermal energy from the heat source by arrows and the movement of butter particles by double lines. Nevertheless, GPT-4V's inference is plausible to some extent because the arrows can work as the sign of the particle's motion as in Figure \ref{Fig:examplePrediction}. This shows that GPT-4V could make incorrect but understandable inferences on the given image and that GPT-4V might have difficulty in processing too contextualized or too sophisticated semantics.

\section {Discussion}\label{}

To our knowledge, this study is one of the first attempts to examine the performance of GPT-4V on multinomial classification tasks, particularly for student-drawn image responses to science items. We developed an approach, NERIF, leveraging prompt engineering to apply GPT-4V to score student drawn models. The results of this study show that GPT-4V processes images and assesses student-drawn models with a varying degree of accuracy. The strength of GPT-4V, which affords visual answer questioning, sharply distinguishes itself from previous approaches in the following aspects.

First, GPT-4V provides a paradigm change in the application of computer vision technologies in educational settings. We found that users can instruct GPT-4V to assess student-drawn models via its powerful image classification only by providing it with \textit{problem context}, \textit{noted rubrics}, and scoring \textit{example}s. That is, GPT-4V requires no programming skills for users in preparing automatic scoring models for visual scientific models. This change is made available due to the prompt engineering approach brought with the progress of AI. In contrast, the previous approaches used by researchers required sophisticated machine learning techniques to train and validate the machine algorithmic models \citep{zhai2022applying, lee2023automated}. Also, while previous reports on GPT-4V provided few-shot learning images in a multi-turn conversation \citep{wu2023early}, this study showed that it is possible to give GPT-4V multiple training examples in a single-turn conversation. This implies that automatic scoring of student-drawn images could be broadly used in educational studies in the near future, tackling the existing technical barrier.

Second, GPT-4V provides interpretative and transparent information that uncovers the "black box" of automatic scoring. In this study, we found that GPT-4V generated answers written in natural language to the modeling tasks, which is understandable to human users and provides rationales for its thoughts on the components defined in \textit{rubric}. This is a significant contribution to automatic scoring, since no automatic image scoring research has provided explainable description of the models. This advantages can help science education researchers to scale up the use of automatic scoring with teachers and students in the near future. The natural language-represented student's image answers and could be a cornerstone that enables timely feedback to students, not only based on the labels of their answers but also the detailed aspects of their drawings described in natural language \citep{zhai2023ai}. The explainability of GPT-4V's scores on automatic grading of image models could be considered more prominent when considering that the explainable AI and scoring model is becoming increasingly important in terms of ethics and transparency of AI in education \citep{khosravi2022explainable}.

Third, we found that GPT-4V requires very few training data on image scoring, which significantly reduced human efforts of labeling training data compared with traditional machine learning approaches. It is notable that previous studies usually sampled a large portion of labeled student responses to train machine. For example, \cite{zhai2022applying, lee2023automated} both sampled around 1,000 student drawn models for each task and hired human experts to score the data, which required a substantial time and cost. In contrast, the automatic scoring of drawn models using GPT-4V in this study only required nine training examples for few-shot learning. This shows that visual language models could also reduce users' burden of data collection that has been mostly used for model training, allowing more data to be used as test cases.

Fourth, this study contributes a new prompting engineering methods (i.e., NERIF) in automatic scoring, which can potentially be generalized for other computer vision tasks. Even though prompt engineering has shown power in many automatic tasks, strategies for specific types of tasks are found essential to improve efficiency. In our case, zero-shot learning approach employed at the beginning of the project showed very low accuracy, which prompted us to explore novel approaches of prompt engineering. We found that the heuristical \textit{instructional notes} supplementary to the scoring rubrics were beneficial in automatic scoring or other educational tasks. This finding is inconsistent with prior research by \cite{chen2023gpt4}, who reported that "supplementing the GPT-4V prompts with reasons underlying the classifications does not yield an improvement in results." We believe that the alignment between the provided reasoning and the model's processing capabilities may differentiated the two studies that yielded contradictory conclusions. In addition, our methods also referred to Chain-of-Thought \citep{wei2022chain, wu2023visual} when developing the 'Rationale for Proficiency.' We also appreciate the ideas provided by \cite{yang2023dawn} and \cite{yang2023set}, which suggested segmentation and marking on the input image to enhance GPT-4V's image understanding. 

Despite the promise of GPT-4V shown in this study, we found limitations. First, our NERIF performed significantly lower with student-drawn models at the 'Proficient' level, as compared to other levels (Table \ref{tab:Confusion_Matrix}). We suspect that this may be because the scoring rubric requires a student's visual answer to include all the components (up to four) to be graded as 'Proficient.' One overly rigorous decision on a component can potentially lead GPT-4V to treat a proficient-like response to be 'Developing.' Future research should develop approaches to mitigating this issue. For example, GPT-4V could be instructed to be less strict and consider as many symbols represented in the image as the indicator of each component. Yet, users should not cautious of risking the GPT-4V to be overly lenient, e.g., failing to differentiating 'Beginning' from 'Developing' answers. Researchers could also test a case for 3 or 5 times and apply some systematic procedures that allow GPT-4V to vote for the ideal solutions, which may increase the accuracy of grading 'Proficient' cases. In all, we suggest that future research should explore more effective ways to instruct GPT-4V to catch and follow the human coders' implicit assumptions working in scoring images need to be sought.

We also find that GPT-4V's conversational function may contaminate the scoring tasks as GPT-4V's memorization of cases intervene its ability to score other cases. As of November 2nd, 2023, the ChatGPT interface allows uploading up to 4 images at once with text prompts. However, when we gave GPT-4V four images at once, it often processed the first two items and ignored the third and fourth images. Therefore, we had to design prompts to use only two images, one for noted rubrics and another for test cases. This issue is expected to be resolved in future developments, especially when the API for GPT-4V is released.

In addition, the scoring accuracy of GPT-4V's reported in this study are not ready to be applied in real classroom settings. Specifically, human-machine scoring agreements indicated by weighted Kappa (.32-.51)(Table \ref{tab:Test_Accuracy})are not comparable to the previous studies which achieved weighted Kappa of .54-.82 \citep{zhai2022applying} or .73-93 \citep{lee2023automated}. Therefore, future studies should explore ways to increase the classification performance of GPT-4V to exploit its full potential for educational studies.

\section {Conclusions}\label{}

This study developed a novel prompt engineering approach--NERIF to leverage GPT-4V in scoring students' scientific models. By testing its image classification performance, this study demonstrated the potential of GPT-4V in scoring student-drawn models . To test the scoring accuracy, we used perfectly balanced data from six assessment tasks in middle school science that include students' drawn models and the scores assigned by human experts. Our findings suggest that GPT-4V could score student drawn-models with low to medium accuracy, varying across student proficiency levels. The study further uncovers how GPT-4V assigned scores in an interpretable way according to NERIF. We found that GPT-4V can retrieve information from input images in terms of the problem context, example evaluations provided by human coders, and students' drawings. GPT-4V catches and describes the characteristics of student-drawn models in natural language and can classify student-drawn models according to the given rubrics. Our approach highlights the few-shot learning with heuristically added "instructional Notes" which improve GPT-4V's performance. In addition, even though GPT-4V made errors, some cases are interpretive to content experts, which indicates space to improve the scoring accuracy. The results of this study show that utilizing GPT-4V in automatic scoring of student-drawn models in science education is promising, leaving gaps to improve the scoring accuracy.

It is expected that OpenAI will soon release the GPT-4V API \citep{hu2023exclusive}, which will enable developers to utilize GPT-4V in more precise, reliable, and efficient ways. The design and development of a prompt that is fed to GPT-4V API with image inputs are expected to resolve the limitations of this study, opening the possibilities to increase image classification accuracy and optimize the memory issue of GPT-4V. Lastly, we would like to highlight that automatic scoring contributes to the validity of assessment uses in education, but in no cases users should rely on solo sources to determine how to use assessment results \citep{zhai2021validity}. Further studies on the ways to apply GPT-4V in education studies, including but not limited to automatic scoring, are strongly recommended.

\section*{Acknowledgement}
This study was funded by the National Science Foundation(NSF) (Award no. 2101104, 2138854). Any opinions, findings, conclusions, or recommendations expressed in this material are those of the author(s) and do not necessarily reflect the views of the NSF. The authors thank the NGSA team and the researchers involved in the parental study (Zhai, He, and Krajcik, 2022) and those who coded the student-drawn models.
The authors specifically thank Joon Kum, who helped develop the prompts and make predictions on the test data.

\bibliography{sn-bibliography}% common bib file

\end{document}